%% file: main.tex
\documentclass[10pt,twocolumn]{article}
\usepackage{graphicx}
\usepackage[table]{xcolor}
\usepackage{ahsmeeting}
\usepackage{amsfonts}
\usepackage{amsmath}
\usepackage{booktabs}
\usepackage{enumitem}
\usepackage{makecell}
\usepackage{subcaption}
\usepackage{url}
\usepackage{caption}
\usepackage[
    colorlinks=true,
    urlcolor=blue,
    linkcolor=blue,
    citecolor=blue
]{hyperref}
\newcommand{\forumyear}{2026}

\meeting{Presented at the Vertical Flight Society's 82nd Annual Forum~\&
\mbox{Technology} Display, West Palm Beach, FL, USA, May 5--7, 2026.
Copyright~\copyright~\forumyear\ by the Vertical Flight Society. All rights
reserved.}

\makeatletter

\newcommand{\suppvideoblock}{%
\vspace{-3.5em}
\begin{center}
\large
\textbf{Video:}
\href{https://youtu.be/mU0zW8kOQzs\&t}
{Click here to watch the video}
\end{center}
\vspace{0.1em}
}

\renewcommand{\maketitle}{%
\par
\begingroup
    \renewcommand\thefootnote{\@fnsymbol\c@footnote}%
    \def\@makefnmark{\rlap{\@textsuperscript{\normalfont\@thefnmark}}}%
    \long\def\@makefntext##1{\parindent 1em\noindent
            {\@textsuperscript{\normalfont\@thefnmark}}##1}%

    \if@twocolumn
      \ifnum \col@number=\@ne
        \@maketitle
        \suppvideoblock
        \@bstract
      \else
        \twocolumn[
            \@maketitle
            \suppvideoblock
            \@bstract
        ]%
      \fi
    \else
      \newpage
      \global\@topnum\z@
      \@maketitle
      \suppvideoblock
      \@bstract
    \fi

 \@thanks
 \@meeting
 \endgroup
 \setcounter{footnote}{0}
}

\makeatother

\begin{document}

\title{Adaptive Outer-Loop Control of Quadrotors via Reinforcement Learning}

\author{\quadauthor{\textbf{Vishnu Saj}\footnote{Corresponding author: vishnu02saj@tamu.edu} \\ Graduate Research Assistant \\ Texas A\&M University \\
College Station, TX, USA}
{\textbf{Sushil Vemuri}\\ Graduate Research Assistant \\  Texas A\&M University \\
College Station, TX, USA}
{\textbf{Dileep Kalathil}\\ Associate Professor \\  Texas A\&M University \\
College Station, TX, USA}
{\textbf{Moble Benedict}\\ Professor \\ Texas A\&M University \\
College Station, TX, USA}
{}}

\date{}

\input{sec_00_abstract}
\maketitle
\input{sec_01_notation}
\input{sec_02_introduction}
\input{sec_03_related_works}
\input{sec_04_preliminaries}
\input{sec_05_methodology}
\input{sec_06_experiments}

\input{sec_07_conclusions}
\input{sec_08_acknowledgements}

\bibliographystyle{ieeetr}
\bibliography{my_refs}

\end{document}

%% file: sec_00_abstract.tex
\abstract{
Deep Reinforcement Learning (DRL) for quadrotor flight control typically relies on Domain Randomization (DR) for sim-to-real transfer, resulting in overly conservative policies that struggle with dynamic disturbances. To overcome this, we propose a novel adaptive control architecture that actively perceives and reacts to instantaneous perturbations. First, we train an optimal outer-loop policy, then replace its reliance on ground-truth disturbance data with a Residual Dynamics Predictor (RDP). The RDP estimates the external forces and moments acting on the aircraft in flight online using only the history of states and control actions. For seamless hardware transfer, we introduce a data-efficient linear calibration bridge and an online thrust correction mechanism that align the simulated latent space with reality using mere seconds of flight data. Real-world validations on a Crazyflie micro-quadrotor demonstrate that our adaptive controller significantly outperforms baselines, maintaining precise trajectory tracking under severe uncertainties including mass variations, asymmetric payloads, and dynamic slung loads.
}

%% file: sec_01_notation.tex
\section{Notation}

\begin{center}
\begin{tabular}{ l l }
    $A, B$ & Spatial amplitudes for trajectory tracking \\
    $\mathcal{A}$ & Reinforcement learning action space \\
    $\mathbf{a}_{t}$ & Action vector at time $t$ \\
    $c_{(\cdot)}$ & Reward function coefficients \\
    $\mathbf{d}_t, \hat{\mathbf{d}}_t$ & Ground-truth and estimated 6D disturbances \\
    $F_{\text{bias}}$ & Accumulated altitude correction thrust bias \\
    $f_g$ & Force due to gravity \\
    $f_{\text{net}}, \tau_{\text{net}}$ & Net forces and moments acting on the quadrotor \\
    $F_T, Q$ & Individual rotor thrust and reaction torque \\
    $H$ & Historical observation window length \\
    $I_{xx}, I_{yy}, I_{zz}$ & Diagonal elements of the inertia tensor \\
    $K_T, K_Q$ & Rotor thrust and torque coefficients \\
    $L_1, L_2$ & Rotor distances from the geometric center \\
    $m$ & Nominal mass of the quadrotor \\
    $N$ & Total number of samples for RMSE calculation \\
    $\mathbf{p}_t, \mathbf{p}_t^{\text{des}}$ & Actual and desired position vectors at time $t$ \\
    $\mathcal{P}$ & Environmental transition dynamics \\
    $\mathbf{q}_t$ & Attitude represented as a quaternion at time $t$ \\
    $r_t$ & Scalar reward at time $t$ \\
    $\mathcal{R}$ & Reinforcement learning reward function \\
    $\mathcal{S}$ & Reinforcement learning state space \\
    $T$ & Time period for dynamic trajectories \\
    $\mathbf{v}_t$ & Linear velocity vector at time $t$ \\
    $\mathbf{x}_t$ & Input feature vector for the RDP \\
    $x_0, y_0$ & Geometric center offset w.r.t. C.G. \\   
\end{tabular}
\end{center}

\begin{center}
\begin{tabular}{ l l }
    $z_{\text{des}}, z_{\text{ref}}$ & Desired / reference flight altitude \\
    $\gamma$ & Reinforcement learning discount factor \\
    $\eta$ & Adaptive integration gain for altitude correction \\
    $\theta$ & Policy parameters / pitch angle \\
    $\pi_{\theta}$ & Stochastic control policy \\
    $\tau_m$ & Motor acceleration/deceleration time constant \\
    $\phi, \theta, \psi$ & Roll, pitch, and yaw Euler angles \\
    $\boldsymbol{\omega}_t$ & Angular velocity vector at time $t$ \\
    $\Omega, \Omega_c$ & Actual and commanded rotor angular speeds \\
\end{tabular}
\end{center}

%% file: sec_02_introduction.tex
\section{Introduction}

Autonomous UAVs are increasingly deployed in complex, real-world applications that demand high levels of agility, precision, and operational reliability, such as infrastructure inspection, search and rescue, and navigation in cluttered environments. Autonomous operation in these scenarios places significant demands on the flight control system. Specifically, the ability to execute precise maneuvers and maintain accurate trajectory tracking under dynamic and uncertain environmental conditions remains a fundamental challenge and is critical for the safety and efficacy of the overall system.

Traditionally, multirotor flight control relies on classical architectures such as Proportional-Integral-Derivative (PID) control, Linear Quadratic Regulators (LQR), or Model Predictive Control (MPC). While these methodologies are highly effective in nominal flight regimes with accurately identified system parameters, they are heavily dependent on explicit mathematical plant models and linearized approximations. Consequently, their tracking performance often degrades rapidly when the vehicle is subjected to unmodeled dynamics, such as severe aerodynamic disturbances, actuator wear, or sudden variations in payload mass. Addressing these uncertainties using classical methods typically requires exhaustive, time-consuming gain scheduling or complex adaptive control laws that are difficult to tune for highly nonlinear flight envelopes.

Recently, deep Reinforcement Learning (RL) has emerged \cite{bauersfeld2021neurobem,mnih2013playing,wang20251000} as a promising alternative for synthesizing nonlinear control policies without the need for explicit analytical modeling. By formulating the control problem as a Markov Decision Process, RL agents can learn complex control mappings directly from simulated experience. However, base RL controllers trained in ideal simulation environments suffer heavily from the simulation-to-reality or ``sim-to-real'' gap; they often fail catastrophically when deployed on physical hardware due to inevitable discrepancies between simulated and real-world dynamics. The most prevalent solution to this is Domain Randomization (DR), wherein the physical parameters of the simulator (e.g., mass, inertia, wind forces) are heavily randomized during training. While DR yields a more robust policy, it forces the network to learn a single, conservative control strategy that maximizes the worst-case performance across all possible randomized environments. This inherently sacrifices agility and optimal performance in nominal conditions, as the controller remains oblivious to the true, instantaneous state of the environment.

This paper seeks to address the  limitations of both baseline and robust RL controllers. Instead of relying on a static, conservative policy, we investigate whether an RL controller can dynamically adapt to its environment, actively altering its behavior in response to the specific disturbances it is experiencing. We pose the following central research question: 

\begin{quote}
    How can a learning-based flight controller dynamically estimate unmodeled physical perturbations from past  states  and use this latent information to maintain optimal tracking performance while seamlessly bridging the sim-to-real gap?
\end{quote}

To answer this, we propose a novel adaptive control architecture. The main contributions of this work are as follows:
\begin{itemize}
    \item \textbf{Adaptive RL Controller:} We develop a reinforcement learning-based outer-loop controller trained alongside an ``oracle'' that has access to ground-truth environmental disturbances, allowing the policy to learn optimal responses to varying external forces and moments.
    \item \textbf{Residual Dynamics Predictor (RDP):} We design a recurrent neural network RDP that acts as a learned state estimator. By observing a brief temporal history of vehicle states and control actions, this RDP implicitly predicts 6D physical perturbations, effectively replacing the Oracle during real-world deployment.
    \item \textbf{Data-Efficient Sim-to-Real Bridge:} We introduce a computationally lightweight linear calibration layer that maps the simulation-trained estimator's latent space to physical reality using a minimal, few-shot real-world dataset, effectively neutralizing steady-state sim-to-real offsets without the need for extensive retraining.
\end{itemize}

\begin{figure}[t]
    \centering
    \includegraphics[width=\columnwidth]{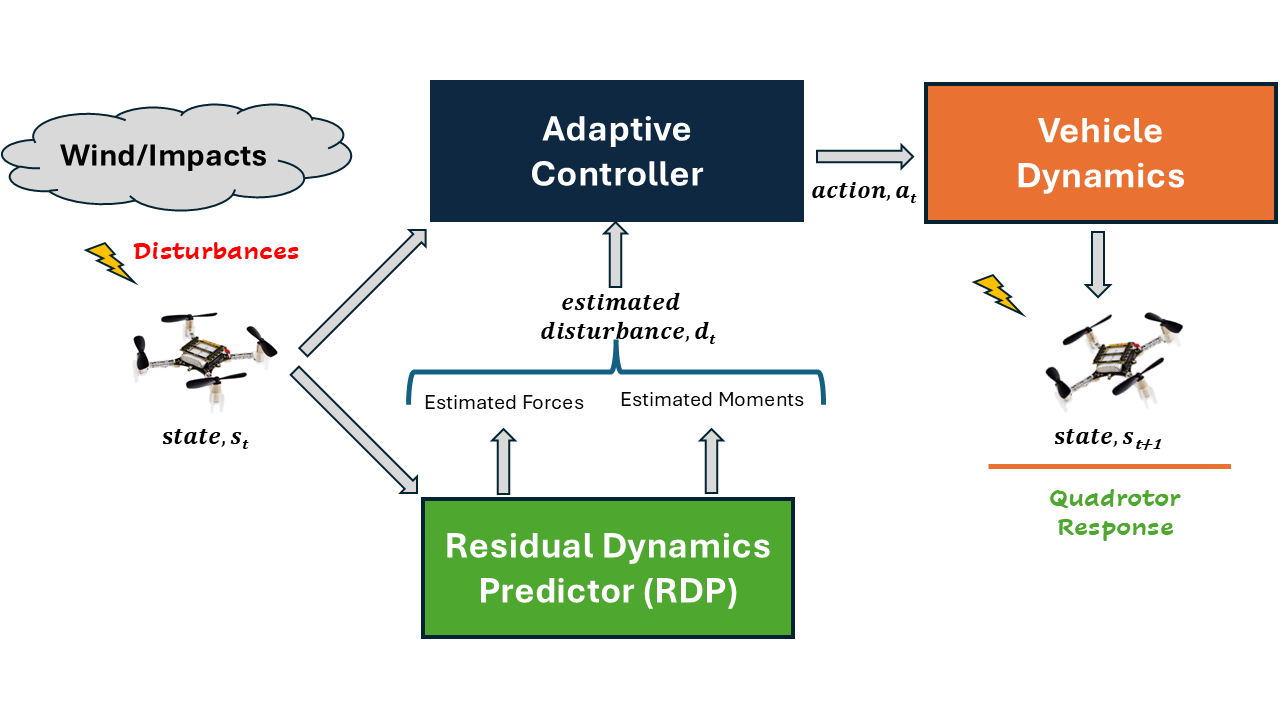}
    \caption{\textbf{Overview of the proposed adaptive control architecture. An ML-based RDP dynamically estimates unmodeled 6D physical perturbations using past states and motor commands, allowing the RL-based Adaptive Controller to react to instantaneous disturbances and seamlessly bridge the sim-to-real gap.}}
    \label{fig: adpative}
\end{figure}

%% file: sec_03_related_works.tex
\section{Related Work}

RL has increasingly been adopted to synthesize flight controllers that can handle the nonlinear dynamics of multirotor vehicles\cite{hwangbo2017control,yoo2020hybrid,batra2022decentralized}. In our previous work \cite{saj2026}, we trained a baseline RL controller and demonstrated its superiority over classical Proportional-Integral-Derivative (PID) architectures in nominal flight conditions. Specifically, the RL-based policy exhibited significantly better tracking performance characterized by lower phase lag and a higher stability margin than the PID baseline. However,  its performance degrades when exposed to out-of-distribution physical perturbations, motivating the need for dynamic adaptation.

To address the limitations of static policies in dynamic environments, recent breakthroughs in legged robotics introduced Rapid Motor Adaptation (RMA) \cite{kumar2021rma}. RMA enables direct deployment to novel environments by using a history of observed states to continuously infer a latent representation of the environment. The adaptive control architecture proposed in this paper is fundamentally inspired by the RMA framework, seeking to translate its success from ground locomotion to the highly dynamic domain of aerial flight.

The concept of rapid adaptation has recently been extended to quadrotors. For instance,  a recent work \cite{zhang2025learning} has successfully implemented an RMA-style architecture for aerial vehicles. However, their methodology relies on abstract latent variables to encapsulate environmental uncertainties, whereas our approach is designed to explicitly estimate the physical 6D external forces and moments acting on the vehicle. Furthermore, their policy is trained end-to-end using a trajectory tracking objective. In contrast, our method relies on a position-hold objective during training, which we found promotes stable, aggressive recovery from severe disturbances while naturally generalizing to dynamic trajectory tracking.

Another notable advancement is RAPTOR \cite{eschmann2025raptor}, which presents a foundation policy for quadrotor control. However, this approach is computationally intensive and complex, requiring the parallel training of hundreds of specialized expert controllers followed by large-scale imitation learning to distill them into a single policy. Similar to \cite{zhang2025learning}, RAPTOR does not explicitly estimate external forces and moments. Conversely, our proposed architecture efficiently trains a single robust policy paired with an explicit, physics-grounded RDP, offering both a computationally streamlined training pipeline and a highly interpretable state estimation mechanism.

%% file: sec_04_preliminaries.tex
\section{Preliminaries and System Modeling}

\subsection{RL Problem Formulation}
We formulate the quadrotor control problem as a Markov Decision Process (MDP) defined by the tuple $(\mathcal{S}, \mathcal{A}, \mathcal{P}, \mathcal{R}, \gamma)$. At each discrete time step $t$, the agent observes the state $\mathbf{s}_t \in \mathcal{S}$, samples an action $\mathbf{a}_t \in \mathcal{A}$ from a stochastic policy $\pi_{\theta}(\mathbf{a}_t | \mathbf{s}_t)$ parameterized by $\theta$, and transitions to a new state $\mathbf{s}_{t+1}$ according to the unknown environmental transition dynamics $\mathcal{P}(\mathbf{s}_{t+1} | \mathbf{s}_t, \mathbf{a}_t)$. The agent subsequently receives a scalar reward $r_t = \mathcal{R}(\mathbf{s}_t, \mathbf{a}_t)$. The objective is to find the optimal policy parameters $\theta^*$ that maximize the expected cumulative discounted reward, $J(\theta) = \mathbb{E}_{\pi_{\theta}} \left[ \sum_{t=0}^{T} \gamma^t r_t \right]$, where $\gamma \in [0, 1)$ is the discount factor and $T$ is the episode horizon.

\subsection{GPU-Accelerated Simulation Environment}
We employ NVIDIA Isaac Lab \cite{mittal2025isaaclab}, a GPU-accelerated robotics framework, for high-performance rigid-body simulation and vectorized training. To achieve precise control over modeling fidelity without altering the core physics engine, we define a custom quadrotor model. Rigid-body kinematics are handled by Isaac Lab, while aerodynamic forces, moments, and actuator dynamics are computed via user-defined external models and applied as external wrenches at each simulation step. We simulate $8,192$ environments in parallel. The physics engine steps at 500 Hz ($\Delta t = 0.002$ s), while the RL policy operates with a control decimation factor of 10, resulting in a control rate of 50 Hz. Training episodes last up to $5.0$ seconds, terminating early if safety boundaries are violated.

\subsection{Quadrotor Dynamics and Actuator Modeling}
The modeled platform is based on the Bitcraze Crazyflie 2.X\cite{crazyflie21} ($m = 0.0315$ kg, $I_{xx}=I_{yy}=1.4 \times 10^{-5} \text{ kg}\cdot\text{m}^2$, $I_{zz}=2.17 \times 10^{-5} \text{ kg}\cdot\text{m}^2$). Operating primarily in near-hover regimes, we assume rotor thrust $F_T$ and torque $Q$ scale quadratically with angular speed $\Omega$:
\begin{equation}
    F_T = K_T \Omega^2, \quad Q = K_Q \Omega^2,
\end{equation}
where $K_T$ and $K_Q$ are experimentally identified coefficients. Individual rotor thrusts are bounded between $0.028$ N and $0.148$ N. 

To bridge the sim-to-real gap, actuator dynamics are modeled as a first-order lag capturing motor and ESC response times:
\begin{equation}
    \dot{\Omega} = \frac{1}{\tau_m}\left(\Omega_c - \Omega\right),
\end{equation}
where $\Omega_c$ is the commanded speed and $\tau_m = 0.03$ s is the time constant.

\begin{figure}[htb!]
    \centering
    \includegraphics[width=0.5\columnwidth]{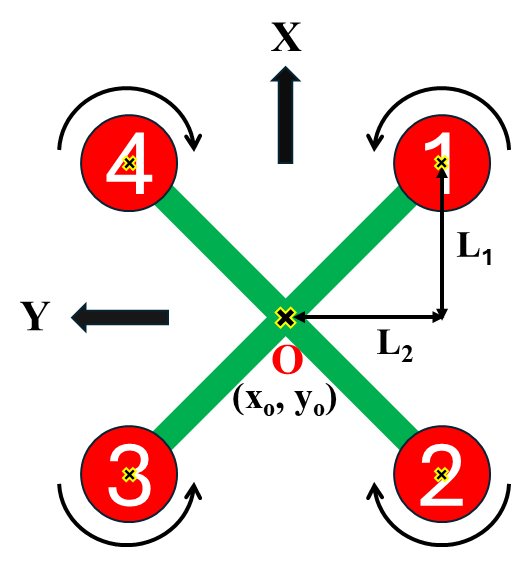}
    \caption{\textbf{Schematic of the quadrotor body frame and rotor configuration. The mathematical model assumes a symmetric cross configuration with explicitly defined center-of-gravity offsets and moment arms, which are utilized to dynamically compute the total aerodynamic forces and moments acting on the vehicle.}}
    \label{fig: quad_pose}
\end{figure}

Assuming a symmetric cross configuration (Fig.~\ref{fig: quad_pose}) with center-of-gravity offsets $(x_0, y_0)$ and moment arms $L_1 = L_2 = 0.028$ m, the total aerodynamic force and moments are:
\begin{equation}
    f_{\text{aero}} =
    \begin{bmatrix}
         0 \\
         0 \\
         K_T\sum_{i=1}^{4} \Omega_i^2
    \end{bmatrix},
\end{equation}
\begin{equation}
\tau_{\text{aero}} =
\begin{aligned}
&
\left[\begin{matrix}
  -K_T(L_1-x_0) & -K_T(L_1+x_0) \\
  K_T(L_2+x_0)  & -K_T(L_2+x_0) \\
  -K_Q          &  K_Q
\end{matrix}\right. \\
&\qquad\qquad
\left.\begin{matrix}
  K_T(L_1-x_0)  &  K_T(L_1+x_0) \\
  K_T(L_2-x_0)  & -K_T(L_2-x_0) \\
  K_Q           & -K_Q
\end{matrix}\right]
\begin{bmatrix}
        \Omega_1^2 \\
        \Omega_2^2 \\
        \Omega_3^2 \\
        \Omega_4^2
\end{bmatrix}.
\end{aligned}
\end{equation}
The net forces and moments acting on the vehicle are obtained by combining these aerodynamic components with gravity ($f_g$).

\subsection{Hardware and Control Architecture}
\begin{figure}[hbt!]
    \centering
    \includegraphics[width=1.0\columnwidth]{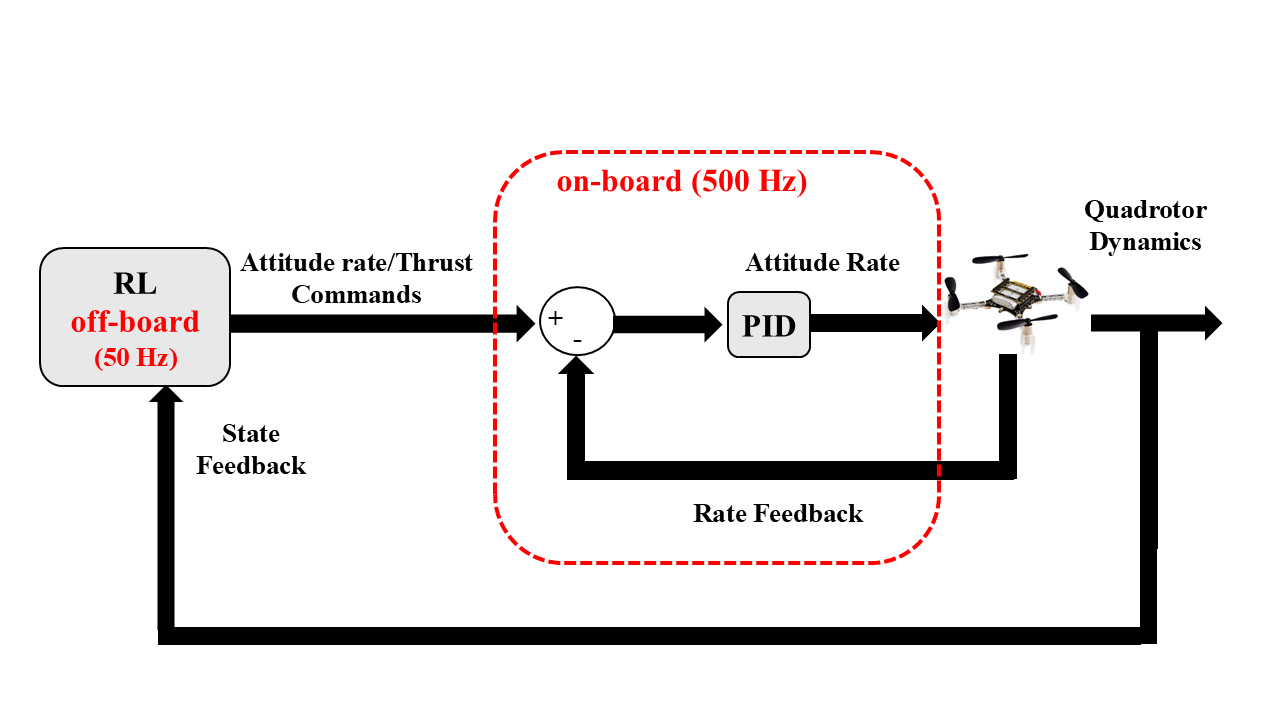}
    \caption{\textbf{Block diagram of the proposed cascaded control architecture. The high-level reinforcement learning policy operates as an offboard outer loop, generating target collective thrust and desired body rates. These setpoints are subsequently transmitted to the vehicle and tracked by the high-bandwidth onboard PID rate controllers.}}
    \label{fig:con_arch}
\end{figure}

We validate our approach on the physical Crazyflie 2.X micro-quadrotor. State estimation relies entirely on the onboard Extended Kalman Filter fusing IMU and Flow Deck (optical flow and time-of-flight) measurements, eliminating the need for external motion capture. We employ a cascaded control architecture (Fig.~\ref{fig:con_arch}). The RL policy acts as an offboard outer-loop controller, outputting desired body rates and collective thrust. These commands are transmitted to the Crazyflie and tracked by its onboard, high-bandwidth PID rate controllers. This modularity isolates the learning-based outer loop while preserving fast and reliable inner-loop stabilization.

\begin{figure}[hbt!]
    \centering
    \includegraphics[width=0.8\columnwidth]{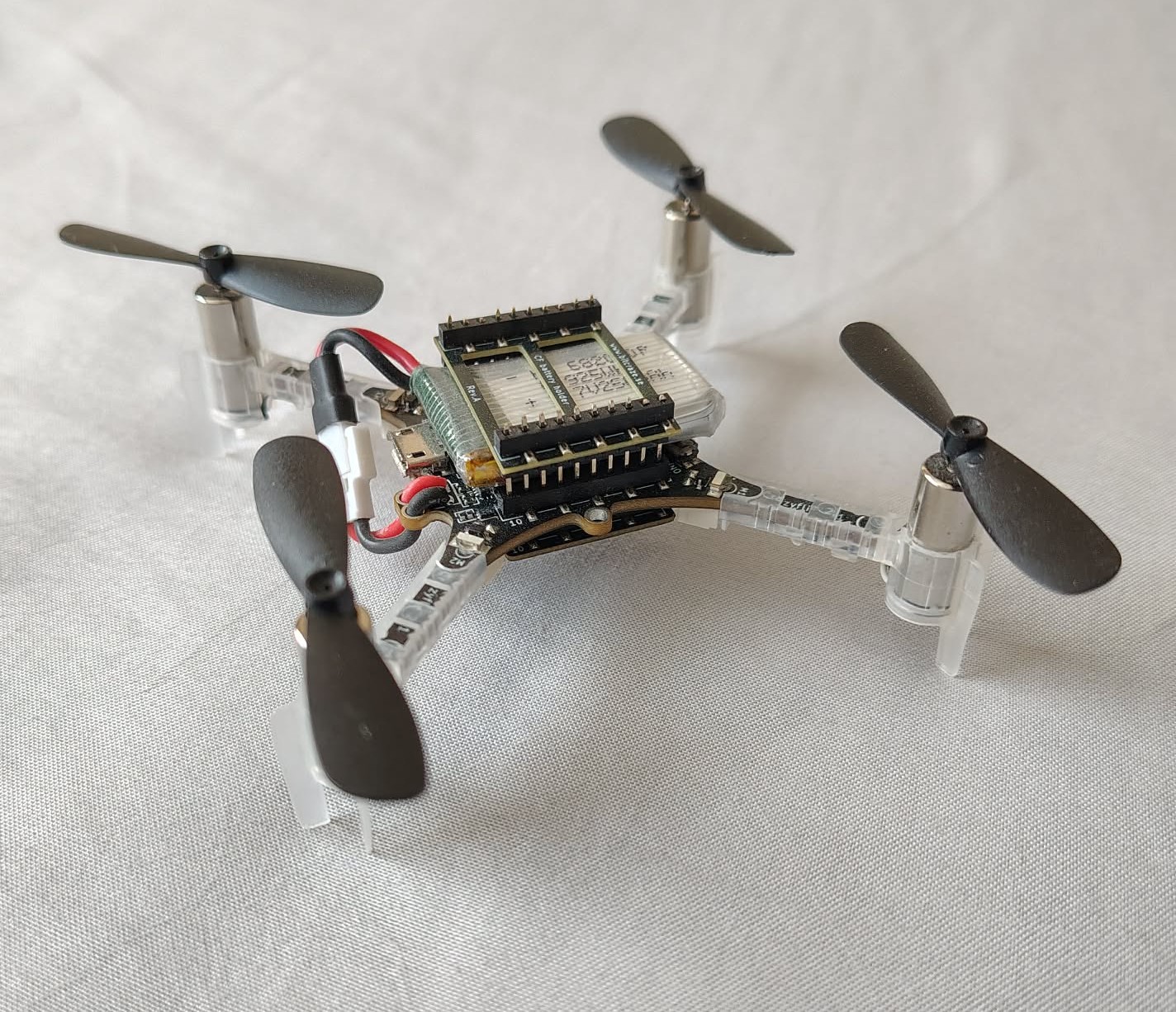}
    \caption{\textbf{The Bitcraze Crazyflie 2.X micro-quadrotor platform utilized for real-world validation experiments. The vehicle relies entirely on its onboard IMU and Flow Deck for precise state estimation, enabling fully autonomous operation without relying on an external motion capture system.}}
    \label{fig:quadcopter}
\end{figure}

\subsection{RL State-Action Space}
The RL policy functions as a high-level outer-loop controller.
\begin{itemize}
    \item \textbf{Action Space ($\mathcal{A} \subset \mathbb{R}^4$):} Collective target thrust and desired body-frame angular rates (roll, pitch, yaw).
    \item \textbf{Observation Space:} The instantaneous observation $\mathbf{o}_t \in \mathbb{R}^{22}$ consists of the vehicle's position error $\mathbf{p}_t \in \mathbb{R}^3$, the flattened rotation matrix $\mathbf{R}_t \in \mathbb{R}^9$ (used to avoid gimbal lock), linear velocity $\mathbf{v}_t \in \mathbb{R}^3$, angular velocity $\boldsymbol{\omega}_t \in \mathbb{R}^3$, and the previous action $\mathbf{a}_{t-1} \in \mathbb{R}^4$. To account for partial observability, such as sensor latency and unmodeled transport delays, the final MDP state $\mathbf{s}_t \in \mathbb{R}^{32 \times 22}$ is constructed by stacking a temporal window of the last 32 observations.
    \item \textbf{Observation Noise:} To simulate real-world estimator drift, zero-mean Gaussian noise is injected into positions ($0.001$ m), orientations ($0.0003$ rad), linear velocities ($0.003$ m/s), and angular velocities ($0.002$ rad/s).
\end{itemize}

\subsection{Reward Formulation}
The objective is to maintain hover at the origin while minimizing control effort. The reward at time $t$ is:
\begin{equation}
\begin{split}
r_t = & \;
c_{p} \|\mathbf{p}_t\|^2 + c_{\psi} \psi_t^2 + c_{v} \|\mathbf{v}_t\|^2 + c_{\omega} \|\boldsymbol{\omega}_t\|^2 \\
& + c_{T} (\Delta T_{t})^2 + c_{\Delta\omega} \|\Delta \boldsymbol{\omega}_{cmd, t}\|^2 + c_{s} + c_{d} \cdot \mathbf{1}_{\text{crash}}
\end{split}
\end{equation}
Coefficients are empirically tuned: position $c_{p} = -30.0$, yaw $c_{\psi} = -2.0$, linear velocity $c_{v} = -3.0$, angular velocity $c_{\omega} = -0.5$, thrust smoothing $c_{T} = -1.0$, and rate smoothing $c_{\Delta\omega} = -0.5$. The agent receives a survival bonus $c_s = 4.0$ and a crash penalty $c_d = -100.0$ (triggered if boundaries $>0.6$ m are exceeded).

\subsection{Network Architecture and Training}
Policies are trained using Proximal Policy Optimization (PPO) \cite{ppo}. To handle partial observability, the policy and value networks are parameterized as Recurrent Neural Networks (RNNs) with two layers and a hidden size of $64$. Training uses a learning rate of $0.001$, $\gamma = 0.99$, and $\lambda = 0.95$, optimized over $8$ epochs and $4$ mini-batches per update. Surrogate clipping and value loss clipping are both $0.2$. Training converges over $100,000$ algorithm timesteps.

\subsection{Baseline Controllers}
To evaluate our adaptive approach, we train two comparative baselines:
\begin{itemize}
    \item \textbf{Base Controller:} Trained under ideal conditions with no external perturbations.
    \item \textbf{Robust Controller:} Trained using Domain Randomization (DR). Random external forces $\mathcal{U}(-0.1, 0.1)$ N and torques $\mathcal{U}(-0.001, 0.001)$ Nm are injected at the center of mass to encourage a conservative, worst-case mitigation strategy.
\end{itemize}

%% file: sec_05_methodology.tex
\section{Methodology}

Having established the foundational reinforcement learning formulation and the baseline robust policy, we now detail our proposed adaptive control architecture. The core objective is to move beyond the conservative, worst-case optimization of Domain Randomization (DR) by enabling the controller to dynamically perceive and react to instantaneous environmental disturbances. This is achieved through a three-stage progression culminating in a learned disturbance estimator and a data-efficient sim-to-real calibration bridge.

\subsection{Adaptive Controller (Oracle Baseline)}
To establish an upper bound on achievable flight performance under uncertain conditions, we first develop an Adaptive Controller trained alongside an ``Oracle''. This policy is trained in the same heavily randomized simulation environment as the Robust Controller, but its observation space is explicitly augmented to include the ground-truth external perturbations. Specifically, the nominal observation vector is expanded to $\mathbb{R}^{32 \times 28}$, appending the 6-dimensional disturbance vector $\mathbf{d}_t \in \mathbb{R}^6$ (comprising three force components and three moment components acting on the vehicle's center of gravity) at each timestep. By operating under the assumption of full state observability, the policy is no longer forced to learn a generalized, conservative compromise. Instead, it learns a specialized control manifold that maps specific dynamic variations to optimal targeted corrective actions. While this oracle controller demonstrates superior tracking performance, it cannot be deployed directly onto physical micro-UAVs, which lack the necessary force-moment sensors to measure 6D aerodynamic disturbances in real-time.

\subsection{Residual Dynamics Predictor (RDP)}
To bridge the gap between the Oracle policy and the limitations in sensing the external forces and moments, we introduce the RDP: a Recurrent Neural Network (RNN) designed to function as a nonlinear, implicit state estimator. This RDP estimates the unobserved 6D environmental perturbations, essentially the external forces and moments, purely from a temporal history of vehicle states and low-level actuation signals. We formulate this system identification problem as a sequence-to-vector regression task. The training dataset comprises approximately 60,000 simulated flight episodes, generated by deploying the pre-trained Oracle controller under randomized disturbances. To accurately infer external forces and moments, the network must observe the residual dynamics of the quadrotor. Thus, the input feature vector $\mathbf{x}_t \in \mathbb{R}^{26}$ consists of the 18-dimensional vehicle kinematic state, the 4-dimensional previous outer-loop action, and crucially, the 4-dimensional Pulse Width Modulation (PWM) signals commanded to the individual motors by the inner-loop PID controller. Including the PWM signals is vital; higher-level thrust and body-rate setpoints fail to uniquely capture transient actuator states, motor lag, and inner-loop saturations necessary for accurate instantaneous force and moment estimation. The architecture employs a Gated Recurrent Unit (GRU)\cite{chung2014empirical} to capture the temporal evolution of the disturbances. It processes a rolling historical window of $H=64$ standardized observations ($\mathbf{X}_{t-H:t}$) through two GRU layers (hidden dimension of 64), followed by a linear readout head that outputs the predicted 6D perturbation vector $\hat{\mathbf{d}}_t$. During real-world inference, the predicted perturbation is smoothed using a rolling buffer of length 32 to mitigate high-frequency prediction noise, and the temporal mean is concatenated with the standard kinematic observation to query the Adaptive Controller.

\subsection{Data-Efficient Sim-to-Real Calibration Bridge}
When transferring the purely simulation-trained RDP to physical hardware (e.g., the Crazyflie platform), a distinct sim-to-real gap inevitably emerges. Unmodeled physical dynamics such as slight mass distribution asymmetries, inherent motor wear, degraded battery voltage curves, and near-surface ground effects cause the GRU to predict a persistent, non-zero offset for external forces and moments, even in a nominal, disturbance-free setting. Rather than undertaking the prohibitively expensive process of collecting large-scale, ground-truth real-world datasets to fine-tune the recurrent network, we introduce a highly data-efficient linear calibration bridge. Utilizing a minimal physical dataset of merely three brief flight samples, we compare the expected nominal states against the GRU's real-world predictions and fit a multivariate linear regression model to the outputs. Crucially, the learned slope of this regression was found to be approximately unity ($m \approx 1.0$), demonstrating that the simulated and physical dynamic manifolds are structurally isomorphic and differ primarily by a constant affine translation. This linear bridge operates as a computationally negligible final layer that effectively neutralizes the steady-state sim-to-real offset. By applying this translation, the latent space of the simulation-trained estimator is precisely aligned with physical reality before it is passed to the RL policy, ensuring robust and accurate real-world flight performance.

Furthermore, to address residual thrust mismatches that may manifest as steady-state altitude errors during physical deployment, we augment the calibration bridge with an online altitude correction mechanism. This mechanism computes an accumulated thrust bias, $F_{\text{bias}}$, based on the real-time altitude tracking error. At each timestep, the bias is updated according to:
\begin{equation}
    F_{\text{bias}} \leftarrow F_{\text{bias}} + \eta (z_{\text{des}} - z),
\end{equation}
where $z_{\text{des}}$ is the desired altitude, $z$ is the current measured altitude, and $\eta$ is a small adaptive integration gain. This accumulated bias is continuously added to the estimator's prediction of the vertical disturbance force, $\hat{F}_z$. By integrating the altitude error over time, this mechanism steadily corrects for unmodeled constant offsets in aerodynamic thrust (such as battery depletion) and evaluates the precise external vertical force prediction required for the policy to maintain the desired altitude.

%% file: sec_06_experiments.tex
\section{Experiments}

To evaluate the efficacy of the proposed adaptive architecture, we conducted a systematic series of real-world experiments. The primary objectives of these experiments are twofold: (1) to quantify the position tracking performance of the adaptive approach compared to baselines under varying disturbance regimes, and (2) to validate the accuracy of the RDP's real-time disturbance estimations against known ground-truth physics.

\subsection{Experimental Setup}

\textbf{Evaluated Controllers:} We compare four distinct control configurations across all experimental scenarios:
\begin{enumerate}[label=(\roman*), leftmargin=*]
    \item \textbf{Base Controller:} Trained in an idealized simulation.
    \item \textbf{Robust Controller:} Trained using Domain Randomization (forces/moments).
    \item \textbf{Adaptive Controller + Oracle:} The theoretical performance upper bound, utilizing ground-truth disturbance information.
    \item \textbf{Adaptive Controller + RDP:} Our fully deployable system, using the GRU-based estimator to infer disturbances online.
\end{enumerate}

\textbf{Evaluation Metrics:} The primary performance metric across all tasks is the Root Mean Square Error (RMSE) of the position tracking error, defined as:
\begin{equation}
\mathrm{RMSE} = \sqrt{\frac{1}{N} \sum_{k=1}^{N} \| \mathbf{p}_k - \mathbf{p}_k^{\text{des}} \|^2},
\end{equation}
where $\mathbf{p}_k \in \mathbb{R}^3$ and $\mathbf{p}_k^{\text{des}} \in \mathbb{R}^3$ denote the actual and desired positions at time step $k$, respectively, and $N$ is the total number of samples. In the tabulated results, external disturbance conditions are quantified by the payload mass expressed as a percentage increase relative to the nominal vehicle mass ($m = 0.0315$ kg).

\textbf{Evaluation Scenarios:} To rigorously stress-test the controllers, we perform three distinct real-world evaluations. We explicitly define the testing brackets for each task to evaluate progressive difficulty:
\begin{itemize}[leftmargin=*]
    \item \textbf{Central Payload Mass Variation:} An additional payload is attached directly to the vehicle's center of gravity. This introduces a constant, known vertical force perturbation without altering the rotational dynamics. In our experiments, the payload mass is explicitly varied across four discrete tiers: 0\%, 7.5\%, 15\%, and 25\% of the nominal quadrotor mass.
    \item \textbf{Asymmetric Payload Distribution:} Mass is attached at the end of one of the quadrotor's arms. This shifts the center of gravity and introduces coupled, constant force and moment perturbations. The payload offsets tested correspond to 0\%, 1\%, 7\%, and 11\% relative mass increases.
    \item \textbf{Dynamic Trajectory Tracking with Suspended Payload:} A mass is suspended from the drone via a tether. The vehicle is commanded to track a planar figure-8 trajectory defined by a Lissajous curve:
    \begin{equation}
        \mathbf{p}^{\text{des}}(t) =
        \begin{bmatrix}
            A\cos\!\left(\frac{2\pi t}{T}\right) \\
            B\sin\!\left(\frac{4\pi t}{T}\right) \\
            z_{\mathrm{ref}}
        \end{bmatrix},
    \end{equation}
    where $A$ and $B$ are the spatial amplitudes, $z_{\mathrm{ref}}$ is the constant target altitude, and $T$ is the total trajectory duration. Varying the period $T$ from $15$ s down to $3$ s scales the required velocities and accelerations. As the quadrotor maneuvers, the suspended payload continuously oscillates, generating complex, unmodeled, and time-varying forces and moments acting on the airframe.
\end{itemize}

\subsection{Experimental Results}

\textbf{Central Payload Mass Variation:} 
Table~\ref{tab:payload_rmse} summarizes the position hold RMSE under varying central payload masses. A key insight from this table is the clear divergence in controller robustness as weight increases. The Base Controller exhibits progressive degradation, dropping from an RMSE of 0.074 m at 0\% to 0.160 m at 25\%. This occurs because its fixed internal dynamics model assumes a constant mass; as the vehicle becomes heavier, the baseline policy outputs insufficient collective thrust, resulting in altitude sag and large positional errors.

Notably, the Robust Controller exhibits the worst performance even in the base scenario (0\% variation), yielding an RMSE of 0.111 m. This poor nominal performance is characterized by severe, high-frequency oscillations. Because the Robust Controller was trained via Domain Randomization to anticipate extreme, randomized 6-DOF disturbances at every timestep, it learns an overly conservative, hyper-reactive policy. In a calm environment, this causes the controller to continuously overcompensate for minor sensor noise, leading to visible jitter. Due to this severe oscillatory instability, the Robust Controller was deemed unsafe for hardware deployment under heavier payload conditions and was not tested further (denoted by `x').

In contrast, the adaptive controller demonstrates consistently strong performance across all configurations, with the variant using the learned RDP outperforming even the oracle-based version (0.024 m to 0.030 m). While this may appear counter-intuitive, it can be explained by considering practical and modeling limitations. The oracle assumes access to exact disturbance information; however, in real experiments, small discrepancies inevitably arise due to factors such as slight variations in payload mass and imperfect placement relative to the center of mass. These introduce subtle errors in the computed "ground-truth" forces and moments. Furthermore, due to sim-to-real mismatches, directly providing these rigid physical quantities to the policy may not align optimally with the dynamics encountered during deployment.

\begin{figure}[hbt!]
    \centering
    \includegraphics[width=1.0\columnwidth]{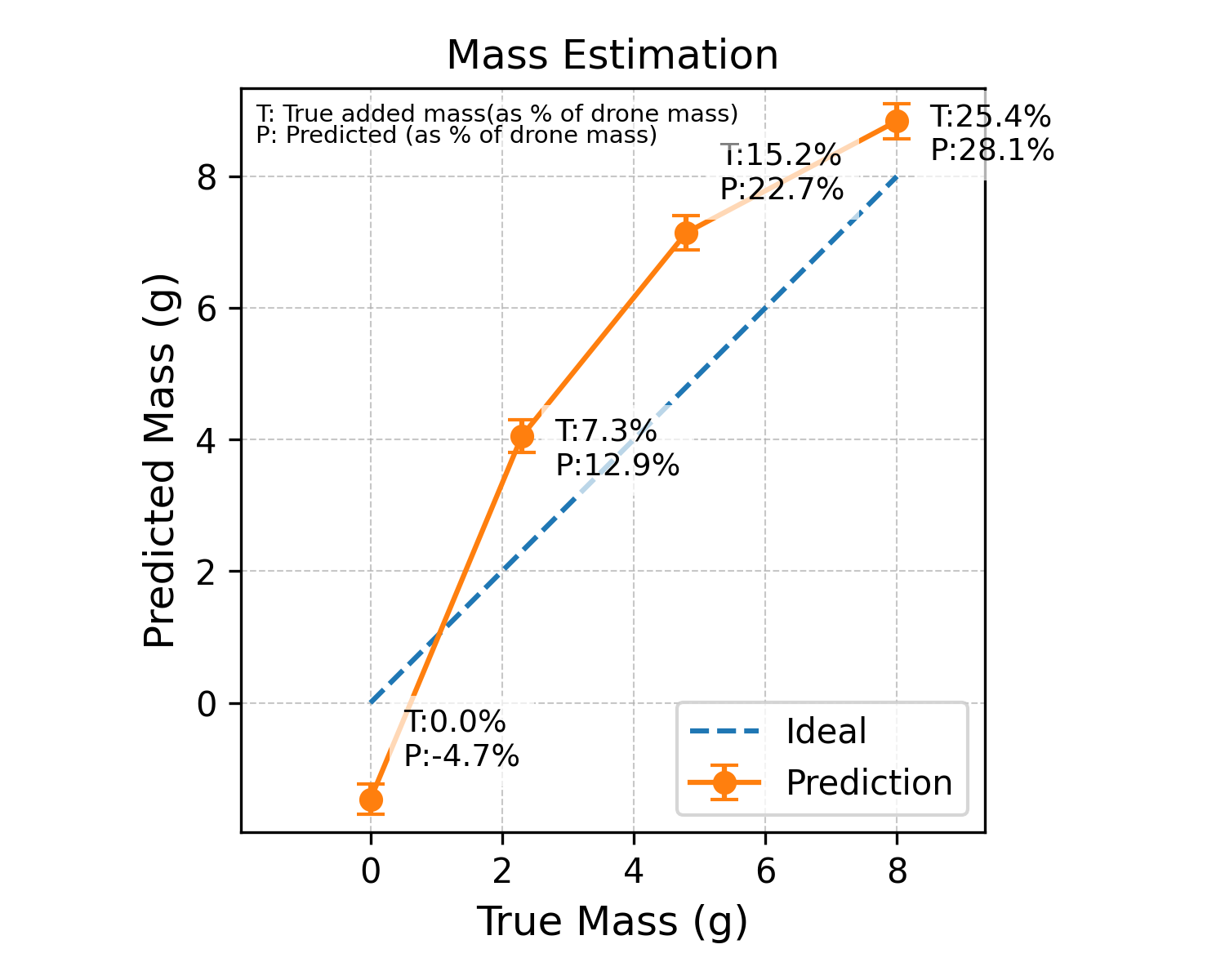}
    \caption{\textbf{Real-time estimation of the added payload mass by the RDP against true mass during the central payload variation experiment . The GRU-based network demonstrates rapid, step-like convergence to the newly altered physical dynamics upon takeoff, maintaining a bounded steady-state estimation error of approximately 3\% to 7\%, which proves sufficient for the RL policy to maintain optimal tracking.}}
    \label{fig:mass_pred}
\end{figure}

The learned RDP, on the other hand, produces a latent disturbance representation that implicitly accounts for such inconsistencies. This behavior is illustrated in Figure~\ref{fig:mass_pred}, where the estimated mass exhibits a bounded error of approximately 3\% to 7\% relative to the true value. Although the estimate does not exactly match the ground truth, it captures the overall trend and relative variations effectively. Importantly, precise disturbance reconstruction is not required; rather, the controller benefits from a consistent and responsive signal that reflects the underlying system behavior. As a result, the RDP provides disturbance information that is better aligned with the actual flight dynamics than idealized oracle inputs.

A similar observation can be made for the baseline controller, where the RMSE varies only marginally for small mass changes (0\% to 7\%), indicating that the system is relatively insensitive to minor deviations from nominal conditions. However, as the mismatch increases, the limitations of fixed dynamics become more pronounced. In this context, the improved performance of the adaptive controller is further supported by the Sim-to-Real Calibration Bridge methodology introduced earlier. By accounting for systematic discrepancies between simulation and real-world dynamics, this approach enables the learned disturbance representation to better align with the true system behavior. Overall, these results suggest that the effectiveness of the adaptive approach stems not from exact physical estimation, but from its ability to provide control-relevant disturbance information while implicitly compensating for modeling mismatches.

\begin{table}[htbp]
    \centering
    \begin{tabular*}{0.9 \columnwidth}{@{\extracolsep{\fill}} lcccc}
        \toprule
        \textbf{Controller} & \textbf{0\%} & \textbf{7.5\%} & \textbf{15\%} & \textbf{25\%} \\
        \midrule
        Base & 0.074 & 0.089 & 0.101 & 0.160 \\
        Robust & 0.111 & x & x & x \\
        \makecell[l]{Adaptive\\ \quad + Oracle} & 0.038 & 0.033 & 0.028 & 0.040 \\
        \makecell[l]{Adaptive\\ \quad + RDP} & 0.024 & 0.028 & 0.029 & 0.030 \\
        \bottomrule
    \end{tabular*}
    \caption{\textbf{Position hold Root Mean Square Error (RMSE) under varying central payload mass configurations. The columns represent the additional payload mass expressed as a percentage of the nominal vehicle mass. The Robust Controller exhibited severe oscillatory instability under nominal conditions and was deemed unsafe for physical deployment under heavier loads (denoted by `x'). In contrast, the Adaptive Controller paired with the learned RDP consistently outperformed the baselines, matching or exceeding the theoretical Oracle performance.}}
    \label{tab:payload_rmse}
\end{table}

\textbf{Asymmetric Payload Distribution:} 
When the payload is offset to the tip of a single arm, the system experiences coupled translational (force) and rotational (torque) disturbances, significantly increasing the task difficulty. As shown in Table~\ref{tab:payload_moment_rmse}, the baseline controller exhibits a gradual increase in tracking error, reaching 0.089 m at 11\% asymmetry. However, the degradation remains relatively moderate compared to the nominal case, indicating that the baseline controller is able to partially accommodate small moment variations despite lacking an explicit mechanism to decouple the induced torques from its thrust commands.

Both the adaptive controller with oracle disturbance information and the adaptive controller with the learned RDP demonstrate significantly improved performance compared to the baseline controller. Their position errors remain consistently lower across all payload configurations, highlighting their ability to effectively compensate for the coupled translational and rotational disturbances. Furthermore, the performance of the adaptive controller with the learned RDP closely matches that of the oracle-based variant, indicating that the RDP is able to provide sufficiently accurate disturbance estimates for effective control.

\begin{figure}[hbt!]
    \centering
    \includegraphics[width=1.0\columnwidth]{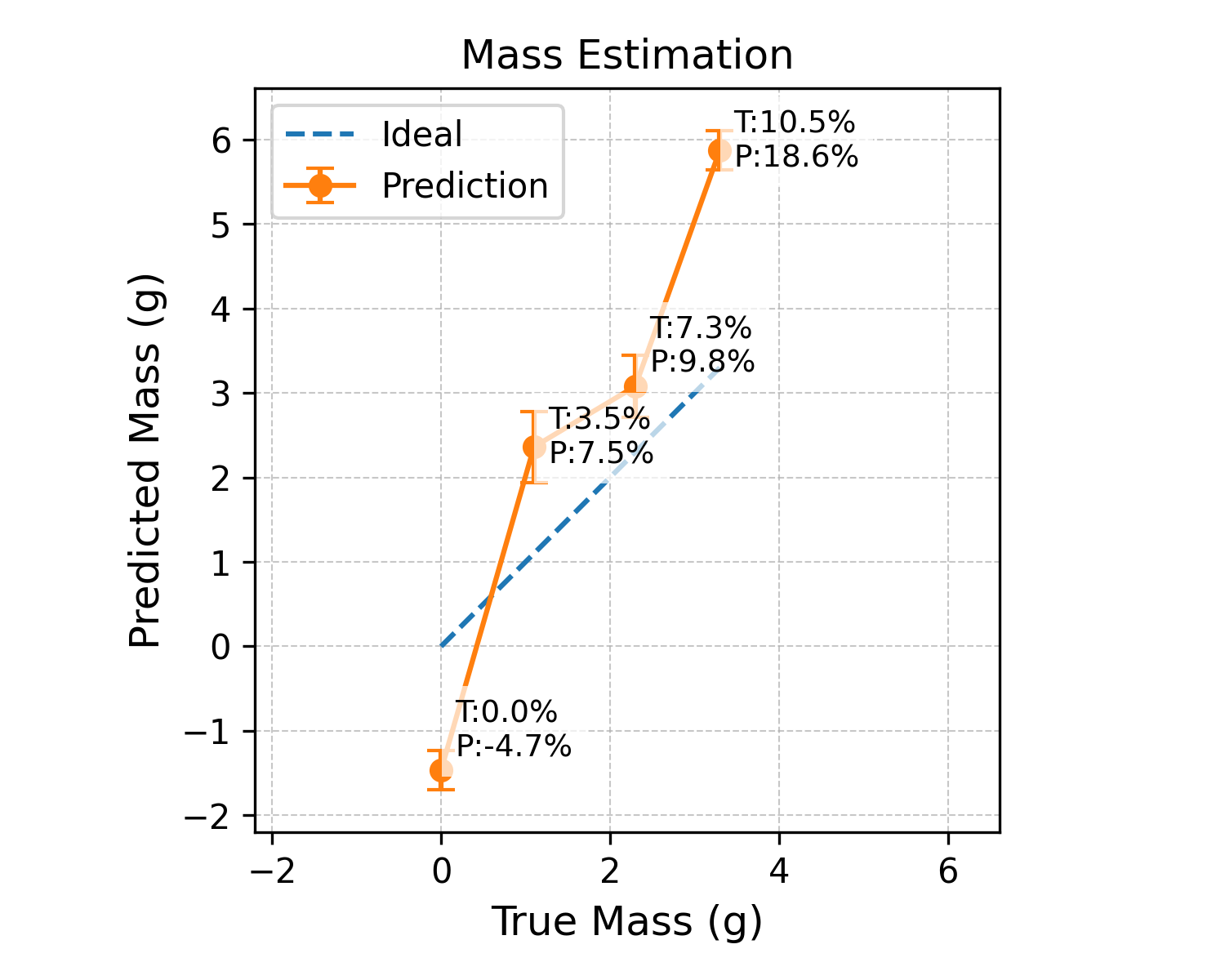}
    \caption{\textbf{Comparison of the additional mass added predicted by the RDP and the true value(also represented as \% of drone mass) under asymmetric loading conditions. Despite the complex coupled dynamics introduced by the offset payload, the vertical force estimation remains stable and accurate.}}
    \label{fig:mass_pred_m}
\end{figure}

\begin{figure}[hbt!]
    \centering
    \includegraphics[width=1.0\columnwidth]{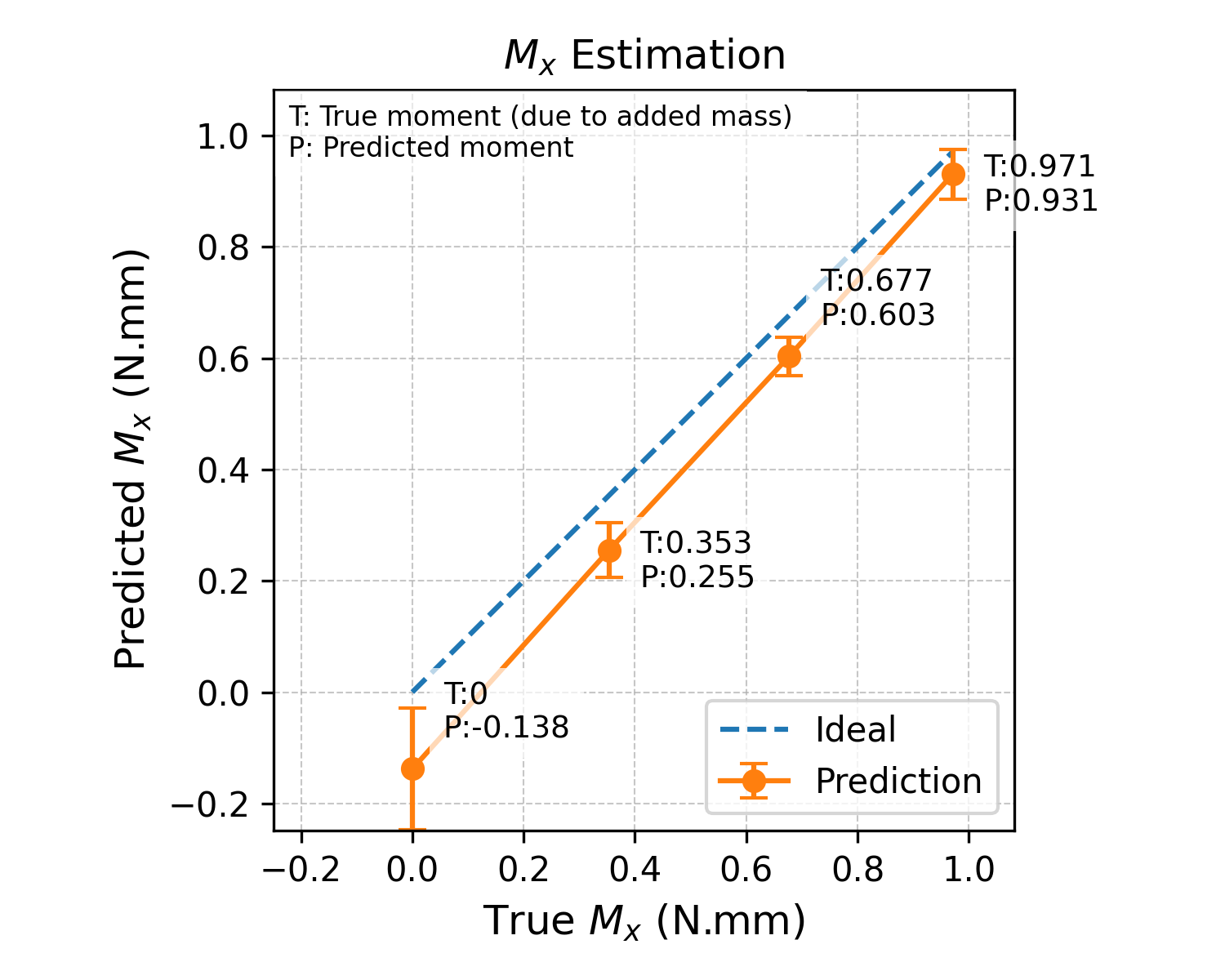}
    \caption{\textbf{Real-time estimation of the induced roll moment ($M_x$) by the RDP during the asymmetric payload experiment. Following a brief transient adaptation phase during takeoff, the RDP accurately converges to the ground-truth steady-state torques utilizing only historical kinematics and motor PWM signals.}}
    \label{fig:Mx}
\end{figure}

\begin{figure}[hbt!]
    \centering
    \includegraphics[width=1.0\columnwidth]{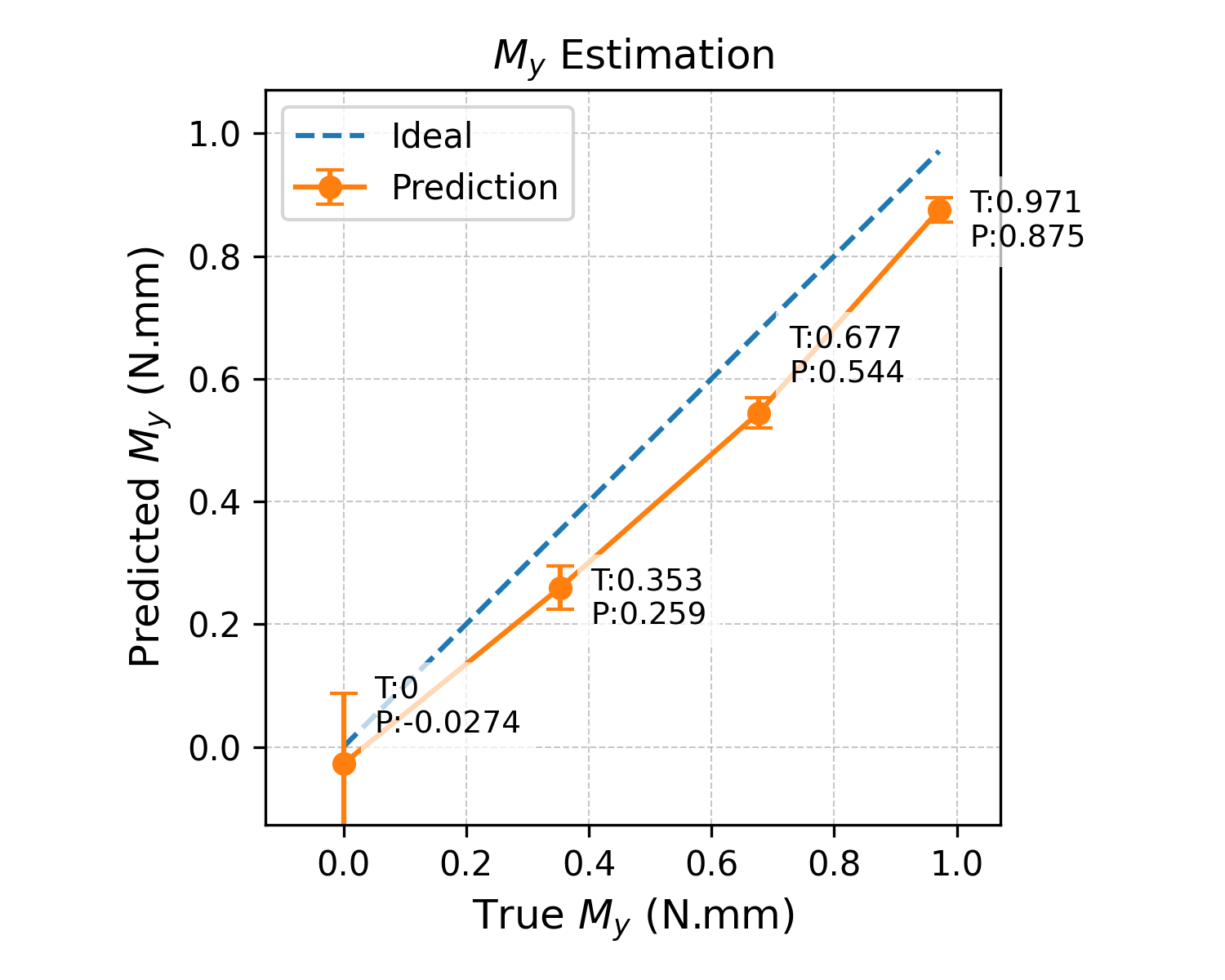}
    \caption{\textbf{Real-time estimation of the induced pitch moment ($M_y$) by the RDP under asymmetric loading. The network successfully isolates the pitch disturbance from the roll and vertical force components, enabling the policy to perfectly counterbalance the offset payload.}}
    \label{fig:My}
\end{figure}

Figures~\ref{fig:mass_pred_m}, \ref{fig:Mx}, and \ref{fig:My} provide crucial insights into how the RDP resolves this complexity. Figure~\ref{fig:mass_pred_m} shows that the total mass estimation remains stable across different configurations. More importantly, Figures~\ref{fig:Mx} and \ref{fig:My} demonstrate that the RDP is able to accurately capture the induced roll ($M_x$) and pitch ($M_y$) moments resulting from the offset center of gravity. The predicted moments closely match the ground-truth values with minimal deviation, indicating high-fidelity estimation of the disturbance-induced torques. This level of accuracy enables the controller to effectively compensate for the coupled rotational effects, contributing to improved overall tracking performance.

A key factor enabling this accurate moment estimation is the inclusion of PWM control inputs in the RDP. These inputs provide implicit information about the thrust distribution across the motors, allowing the RDP to infer the resulting moments acting on the system. Under the assumption of identical motor characteristics, the mapping from PWM signals to generated forces and moments remains consistent, leading to reliable disturbance estimation. However, in practical scenarios, motor characteristics may diverge over time due to wear, manufacturing variability, or partial degradation, introducing asymmetry in thrust generation. In such cases, PWM signals alone may no longer serve as a sufficient indicator of the true forces and moments, potentially leading to degraded estimation accuracy. This highlights a limitation of the current formulation and suggests that incorporating additional information, such as motor-specific calibration or online identification, could further improve the robustness of the RDP under long-term real-world operation.

\begin{table}[htbp]
    \centering
    \begin{tabular*}{0.9 \columnwidth}{@{\extracolsep{\fill}} lcccc}
        \toprule
        \textbf{Controller} & \textbf{0\%} & \textbf{4\%} & \textbf{7\%} & \textbf{11\%} \\
        \midrule
        Base & 0.074 & 0.080 & 0.082 & 0.089 \\
        Robust & 0.111 & x & x & x \\
        \makecell[l]{Adaptive \\ \quad + Oracle} & 0.038 & 0.037 & 0.039 & 0.027 \\
        \makecell[l]{Adaptive \\ \quad + RDP} & 0.024 & 0.033 & 0.026 & 0.039 \\
        \bottomrule
    \end{tabular*}
    \caption{\textbf{Position hold RMSE under asymmetric payload configurations, where mass is attached to a single rotor arm to induce coupled translational and rotational disturbances. Columns indicate the percentage of mass increase. The Adaptive Controller with the learned RDP demonstrates superior robustness, effectively mitigating the induced roll and pitch moments, whereas the Base Controller's performance steadily degrades as the asymmetry increases.}}
    \label{tab:payload_moment_rmse}
\end{table}

\textbf{Dynamic Trajectory Tracking with Suspended Payload:}

\begin{figure*}[t]
    \centering
    
    \begin{subfigure}[b]{1.0\columnwidth}
        \centering
        \includegraphics[width=\linewidth]{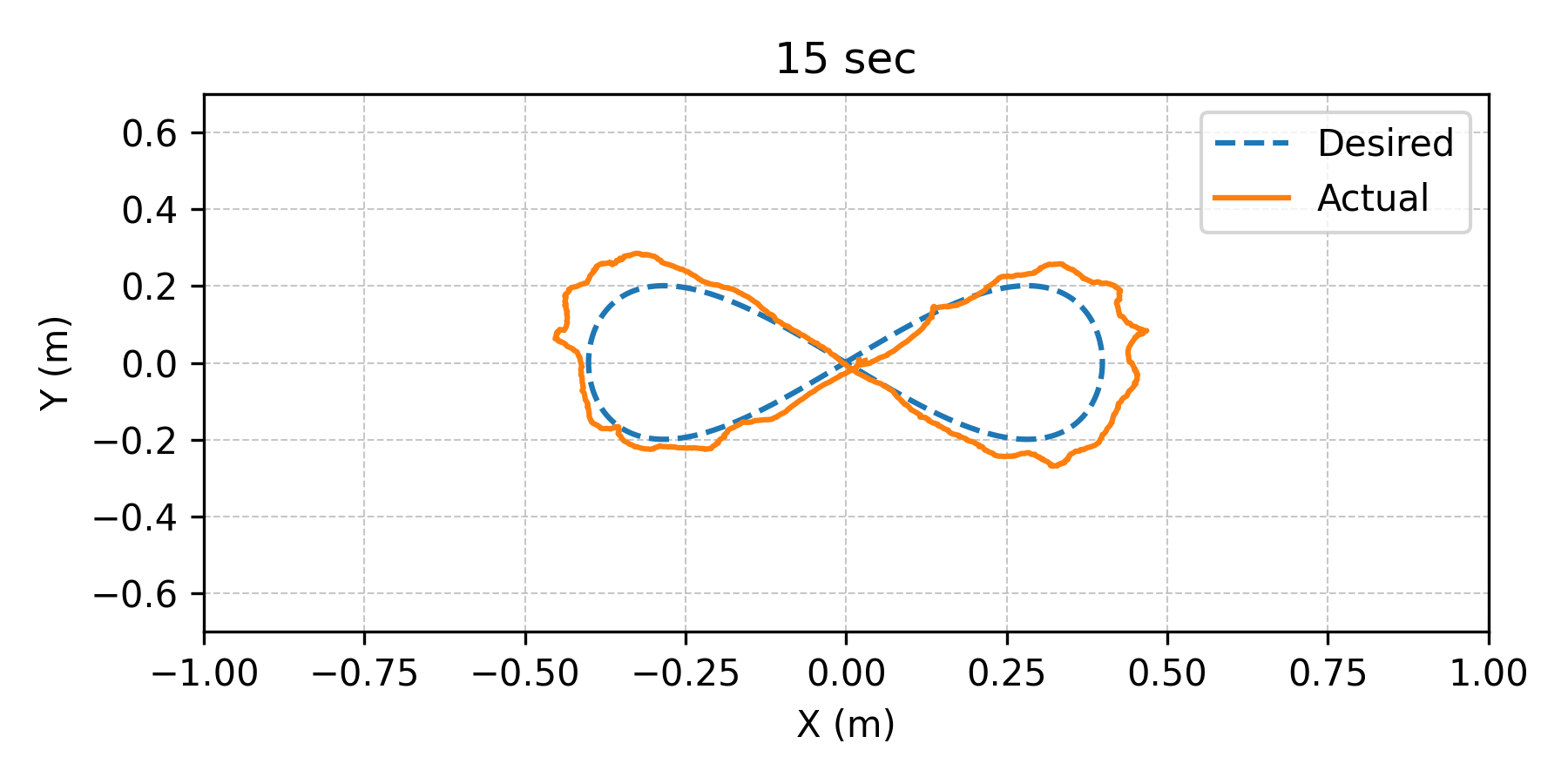}
    \end{subfigure}
    \hfill
    \begin{subfigure}[b]{1.0\columnwidth}
        \centering
        \includegraphics[width=\linewidth]{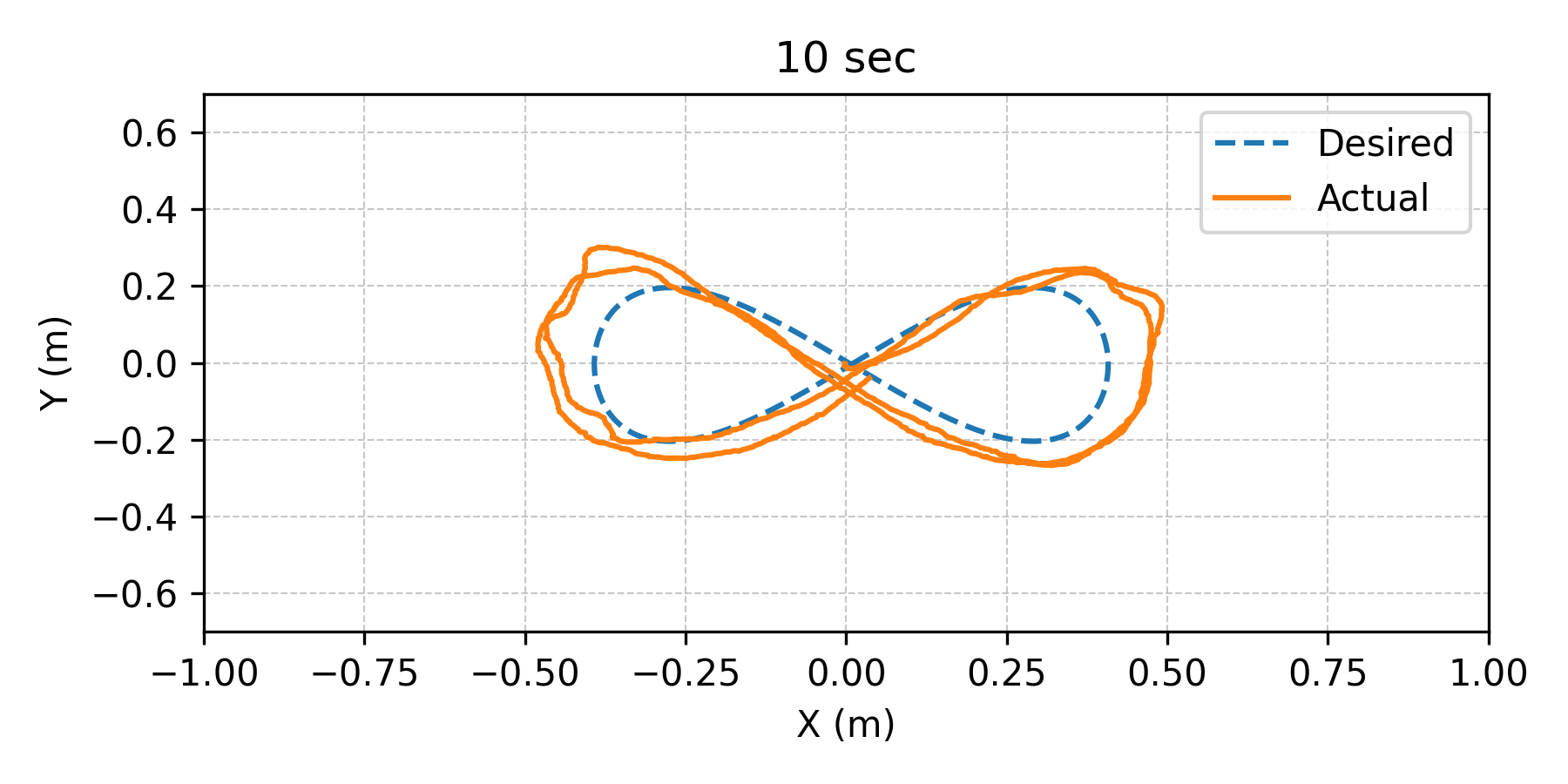}
    \end{subfigure}
    
    \vspace{0.5em}
    
    \begin{subfigure}[b]{1.0\columnwidth}
        \centering
        \includegraphics[width=\linewidth]{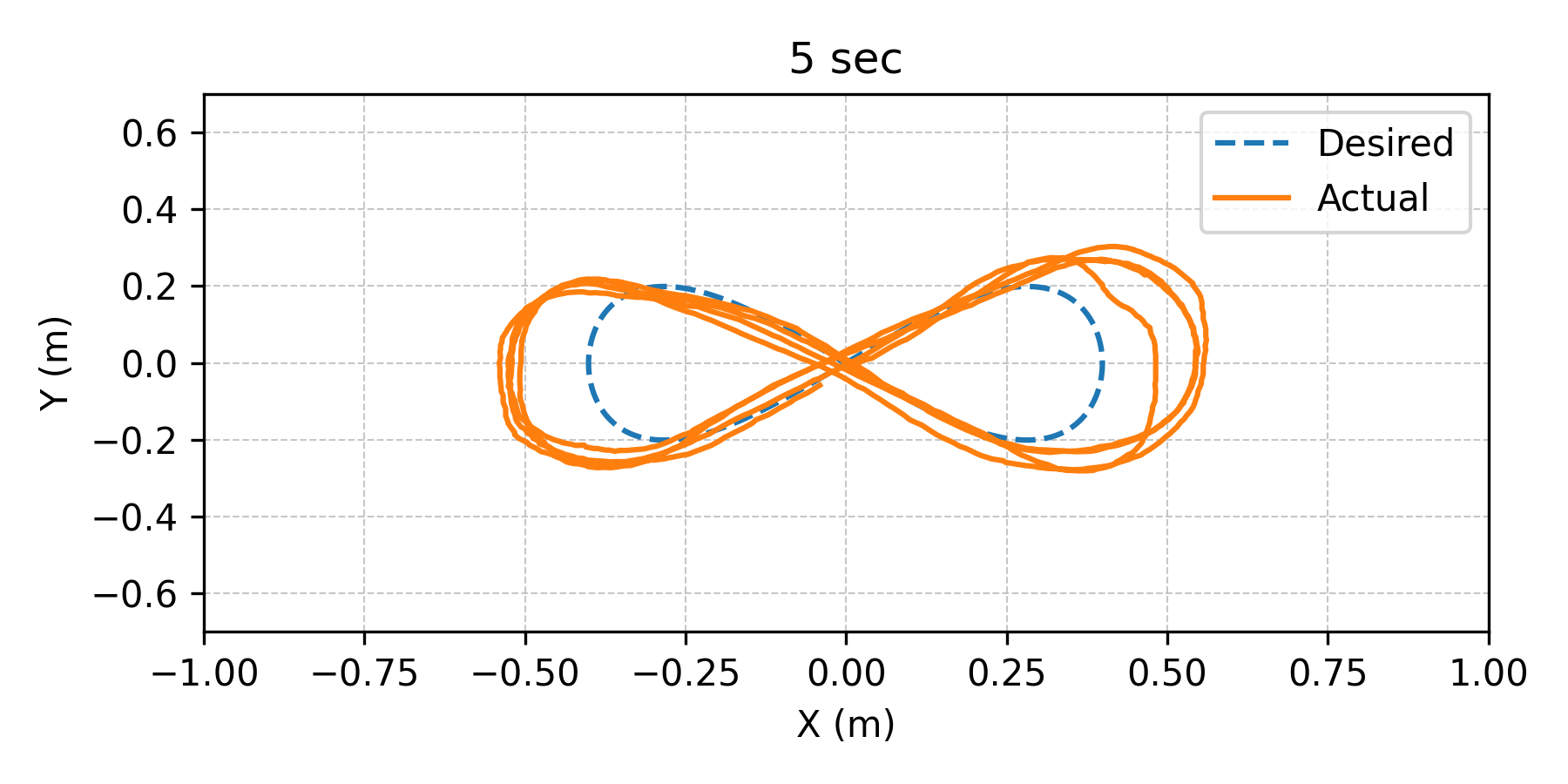}
    \end{subfigure}
    \hfill
    \begin{subfigure}[b]{1.0\columnwidth}
        \centering
        \includegraphics[width=\linewidth]{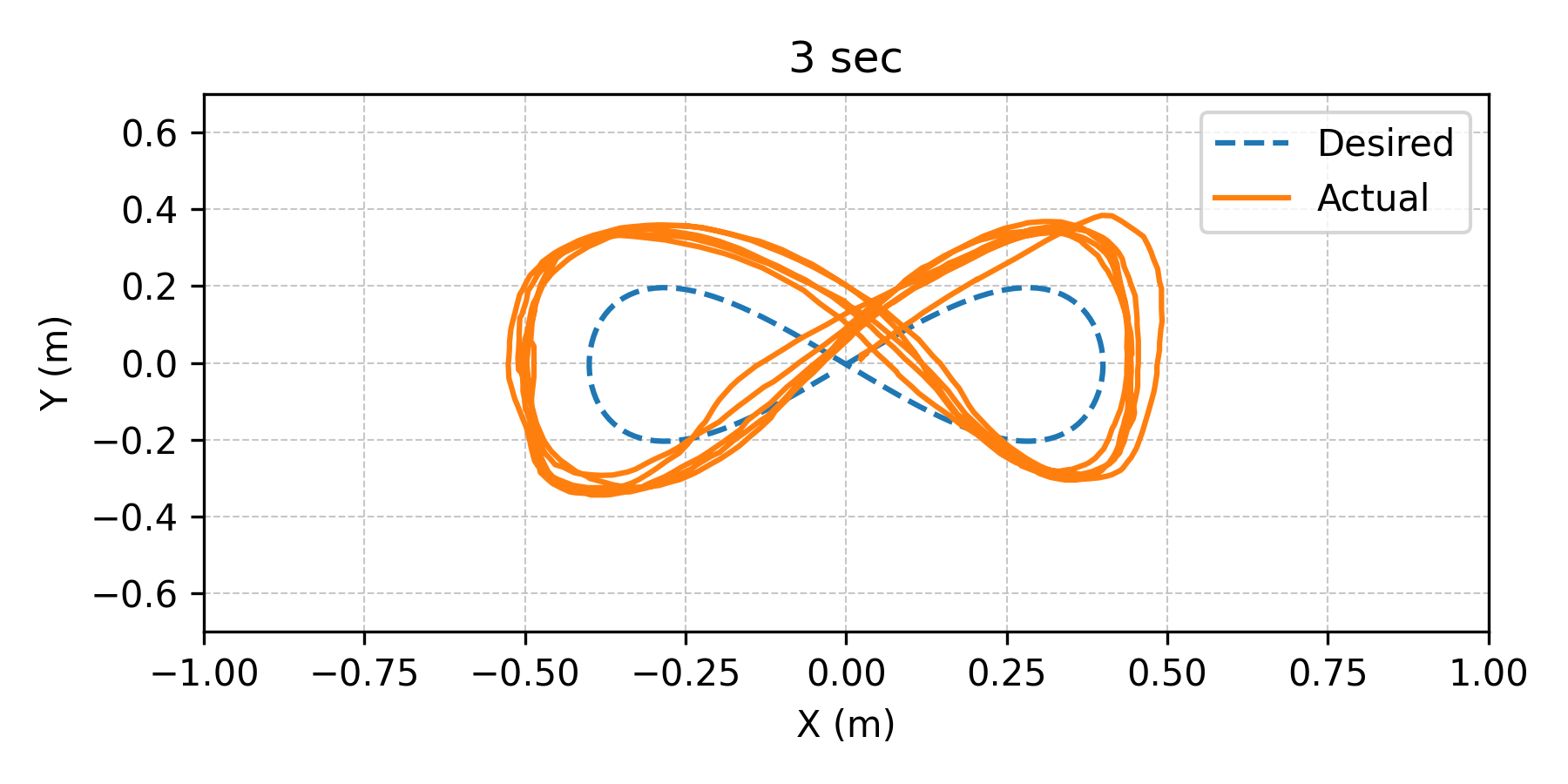}
    \end{subfigure}
    
    \caption{\textbf{Planar $x$--$y$ trajectory tracking performance of the adaptive controller carrying a 4.7 g suspended payload attached via a thread of length equal to the arm length. The reference trajectory is a Lissajous figure-8 curve with decreasing time periods ($T$). As $T$ decreases from 15 s to 3 s, the required velocities and accelerations increase significantly, inducing aggressive pendulum dynamics. Remarkably, the controller The controller maintains stable flight and continues to follow the reference trajectory despite the presence of significant unmodeled oscillatory disturbances.}}
    \label{fig:traj_tracking}
\end{figure*}

\begin{figure*}[h!]
    \centering
    
    \begin{subfigure}[b]{1.0\columnwidth}
        \centering
        \includegraphics[width=\linewidth]{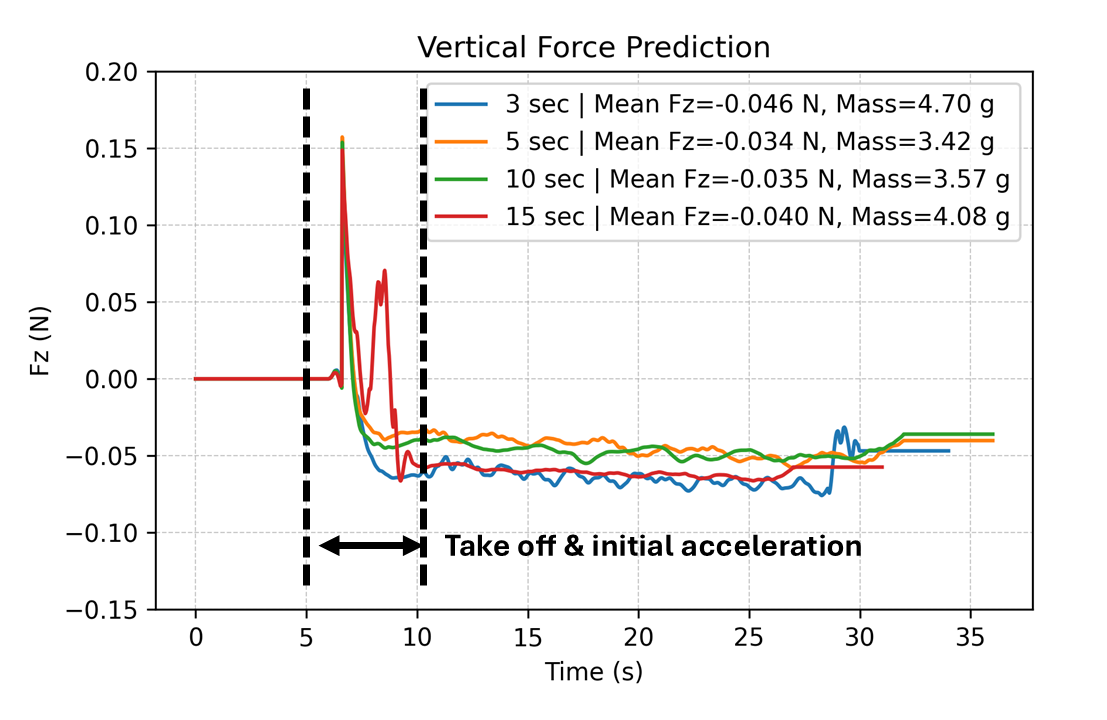}
    \end{subfigure}
    \hfill
    \begin{subfigure}[b]{1.0\columnwidth}
        \centering
        \includegraphics[width=\linewidth]{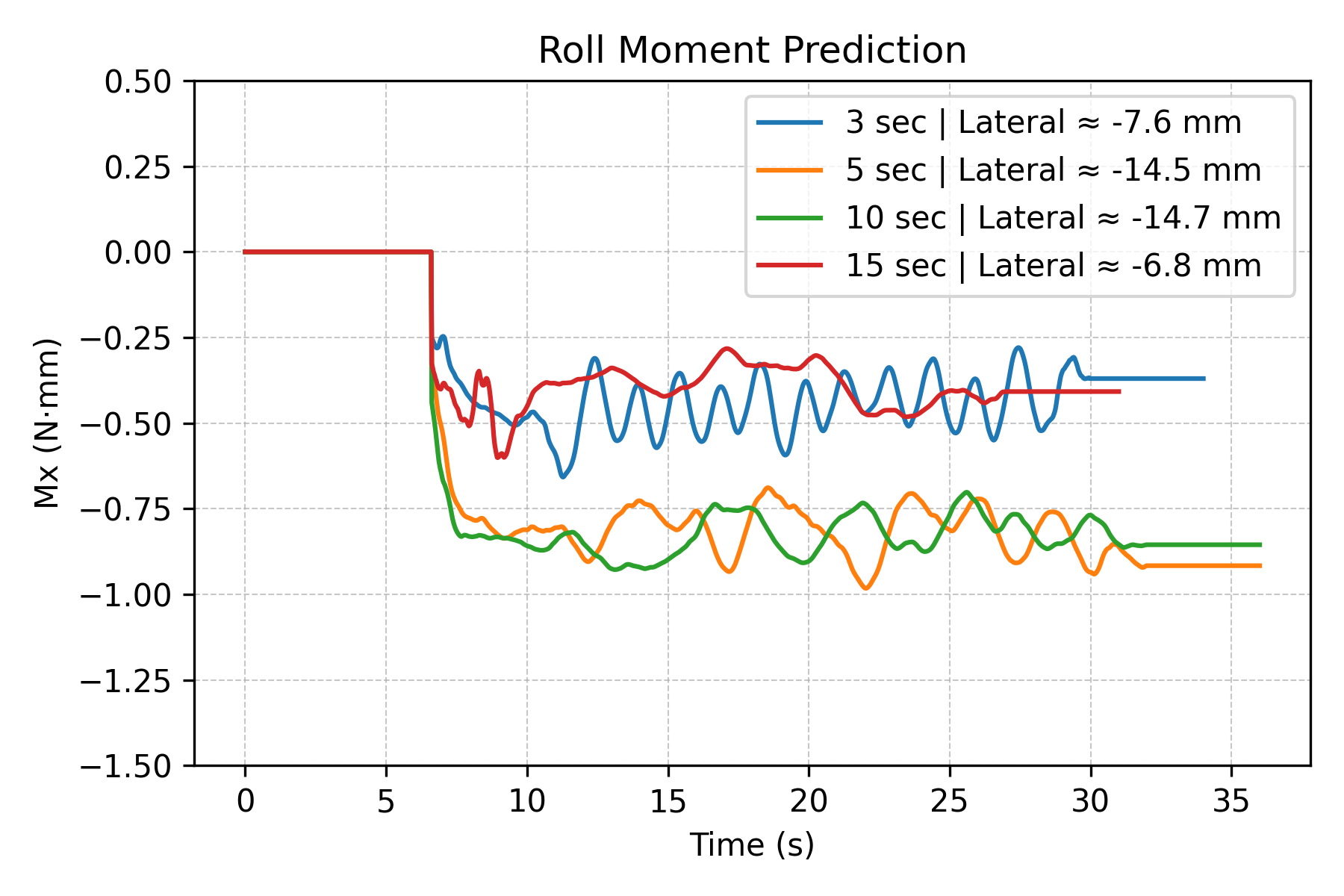}
    \end{subfigure}
    \hfill
    \begin{subfigure}[b]{1.0\columnwidth}
        \centering
        \includegraphics[width=\linewidth]{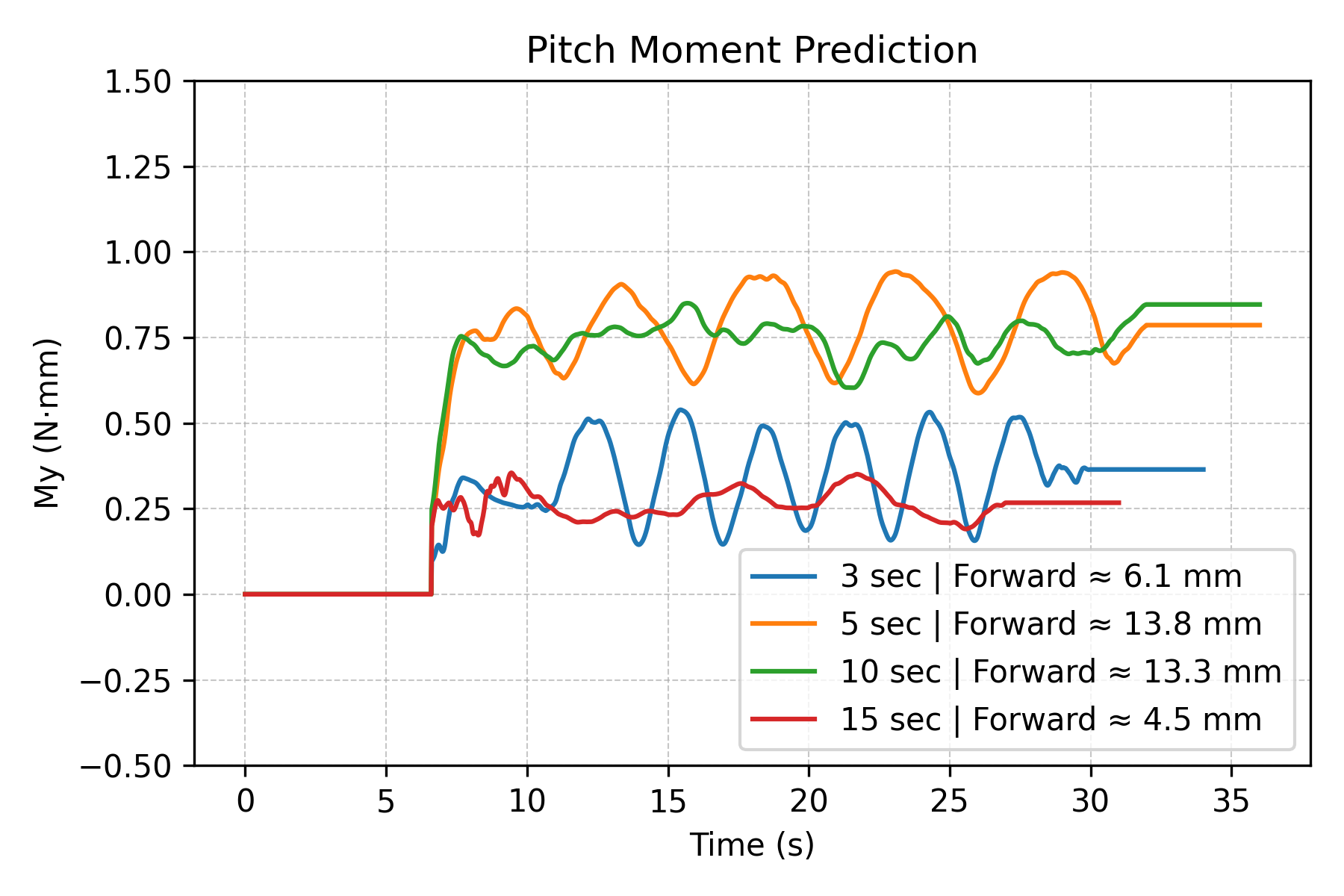}
    \end{subfigure}
    
    \caption{\textbf{Time-series of the estimated disturbance quantities—vertical force ($F_z$), roll moment ($M_x$), and pitch moment ($M_y$)—during the dynamic slung load trajectory tracking experiments. The highly oscillatory predictions actively capture the real-time pendulum physics of the swinging payload, dynamically scaling in amplitude as the trajectory period $T$ decreases and the maneuver aggressiveness increases.}}
    \label{fig:disturbance_traj}
\end{figure*}

For dynamic tracking, a $4.7\,\text{g}$ ($\sim15\%$) payload suspended by a thread of length equal to the arm length introduces continuous, oscillatory disturbances. Figure~\ref{fig:traj_tracking} shows the planar $x$--$y$ trajectories of the adaptive controller tracking the figure-8 reference across decreasing time periods ($T$). As $T$ decreases from 15 s to 3 s, the required velocities and accelerations increase significantly, resulting in more aggressive motion. This, in turn, induces larger oscillations of the suspended payload, generating stronger and more rapidly varying disturbance forces and moments acting on the vehicle.

\begin{table}[htbp]
\centering
\small
\setlength{\tabcolsep}{4pt} 

\begin{tabular}{lccccc}
\toprule
$T$ (s) & RMSE & Max Err & Std & Delay & Phase \\
        & (m)  & (m)     & (m) & (s)   & (deg) \\
\midrule
15 & 0.055 & 0.108 & 0.024 & -0.140 & -3.4 \\
10 & 0.063 & 0.112 & 0.023 & -0.080 & -2.9 \\
5 & 0.118 & 0.192 & 0.041 & 0.080 & 5.8 \\
3 & 0.141 & 0.220 & 0.036 & 0.100 & 12.0 \\
\bottomrule
\end{tabular}
\caption{\textbf{Trajectory tracking performance under increasing aggressiveness. As the trajectory period $T$ decreases, tracking error increases, accompanied by growing time delay and phase lag (computed along the $x$-axis), indicating bandwidth-limited response.}}
\label{tab:traj_tracking}
\end{table}

Table~\ref{tab:traj_tracking} quantifies the effect of increasing trajectory aggressiveness on tracking performance. As $T$ decreases from 15 s to 3 s, the RMSE increases from 0.055 m to 0.141 m, accompanied by a rise in the standard deviation of the error, indicating a more oscillatory response under faster trajectory execution. This trend is further explained by the observed variation in tracking delay and phase. The delay and phase were computed along the $x$-axis, where the dominant motion of the trajectory makes tracking errors most pronounced. For slower trajectories ($T=15$ s and $10$ s), the delay is small and slightly negative ($-0.14$ s and $-0.08$ s), corresponding to a slight phase lead. This behavior suggests that the learned policy is able to implicitly anticipate the system dynamics based on recent state information, effectively exhibiting predictive characteristics. As the trajectory becomes more aggressive, this lead diminishes and transitions into a lag, with the delay increasing to $0.10$ s for $T=3$ s and a corresponding phase lag of $12^\circ$. This indicates that the system increasingly lags behind the reference as the excitation frequency rises. Despite this increase in tracking error, the maximum deviation remains bounded within a limited range (0.108 m to 0.220 m), and no divergence is observed across all experiments. This indicates that the system maintains stable closed-loop behavior even under highly dynamic conditions. Overall, the observed performance trends are consistent with bandwidth limitations of the closed-loop system, where higher-frequency inputs reduce the controller’s ability to maintain anticipatory behavior and track the reference with the same fidelity as slower trajectories.

The underlying mechanism enabling this robustness is illustrated in Figure~\ref{fig:disturbance_traj}. The estimated disturbance quantities, including the vertical force ($F_z$) and the induced roll ($M_x$) and pitch ($M_y$) moments, exhibit strongly time-varying behavior that reflects the oscillatory dynamics of the suspended payload. As the trajectory becomes more aggressive (lower $T$), both the frequency and amplitude of these oscillations increase, indicating stronger and more rapidly varying disturbance effects due to intensified payload motion.

Throughout most of the trajectory, the vertical force estimate—and the corresponding inferred payload mass—remains relatively consistent, suggesting that the RDP is able to maintain a stable representation of the overall load. On average, the RDP predicts a mean mass in the range of 3.4--4.7 g, underestimating the true payload mass of 4.7 g, which is consistent with the estimation error observed in the previous experiment. However, transient inaccuracies are observed during takeoff and at the initial phase of trajectory execution, where the disturbance estimates deviate from their steady-state values. This behavior is expected, as the system undergoes rapid changes in dynamics and the RDP requires a short period to converge. Once the trajectory enters steady-state motion, the vertical force estimates stabilize and provide a reliable representation of the disturbance.

In contrast, the estimated moments oscillate about a mean value, capturing the time-varying moments induced by the swinging payload. Interestingly, two distinct mean levels are observed in the moment estimates across different trials, corresponding to effective payload offsets of approximately 14 mm and 7 mm from the center. This indicates that the RDP is able to infer not only the magnitude of the disturbance but also variations in the effective location of the payload. This behavior can be attributed to the physical interaction between the suspended payload and the vehicle structure, where the attachment point is free to shift slightly along the arm. Due to minor geometric features or contact effects, the payload tends to settle at different effective positions, resulting in distinct moment offsets. Despite this variability, the RDP successfully captures these changes, allowing the controller to account for both the oscillatory and quasi-static components of the disturbance, thereby maintaining stable and consistent tracking performance.

Despite these transient and configuration-dependent variations, the RDP successfully captures both the oscillatory and quasi-static components of the disturbance, enabling the controller to maintain stable and consistent tracking performance.

%% file: sec_07_conclusions.tex
\section{Conclusions and Future Work}

This work presented a learning-based adaptive control architecture for micro-quadrotors operating under severe dynamic uncertainties. By transitioning from conservative domain randomization to an active disturbance-rejection framework, this research yields several key insights into bridging the sim-to-real gap:

\begin{itemize}[leftmargin=*]
    \item \textbf{Active Adaptation Outperforms Passive Robustness:} We demonstrated that relying exclusively on Domain Randomization (DR) to handle unmodeled perturbations forces the policy to learn overly conservative, hyper-reactive behaviors. On physical hardware, this passive robustness translates to severe oscillatory instability even in nominal conditions. Active, real-time adaptation is fundamentally required to maintain stable and optimal flight tracking.
    
    \item \textbf{Implicit Disturbance Estimation from Onboard Signals:} Our experiments validate that complex 6D external forces and moments can be accurately reconstructed without specialized force-torque sensors. A brief temporal history of standard onboard kinematics coupled with low-level actuator signals (PWM) contains sufficient implicit information for a recurrent network (the RDP) to isolate and estimate independent translational and rotational disturbances.

    \item \textbf{Control-Relevant vs. Exact Physical Estimation:} Results from the payload variation experiments indicate that the RDP does not need to perfectly reconstruct the ground-truth physical quantities to be highly effective. A bounded estimation error (e.g. the observed 3\% to 7\% mass error) is entirely permissible. As long as the network generates a consistent and control-relevant representation of the disturbance trends, the RL policy can appropriately scale its compensatory actions to maintain high-precision tracking.
    
    \item \textbf{Data-Efficient Sim-to-Real Calibration:} A key finding of this work is that the discrepancy between simulated disturbance predictions and real-world behavior can be well-approximated by a low-order (approximately affine) relationship. As a result, bridging the sim-to-real gap does not require complex nonlinear domain adaptation or extensive real-world retraining. Instead, it can be effectively mitigated using a computationally lightweight linear calibration bridge combined with an online thrust accumulator, requiring only a few seconds of flight data to correct for steady-state offsets (e.g., battery voltage variations).
    
    \item \textbf{Efficacy of Decoupled, Oracle-Guided Training:} Training an optimal outer-loop policy using a ground-truth ``Oracle'' before transitioning to a learned estimator circumvents the instability and sample inefficiency often associated with end-to-end RL in partially observable environments. This decoupled approach cleanly separates the control objective from the state-estimation problem, ensuring that the policy learns an optimal disturbance-response manifold. This modularity was crucial for achieving aggressive real-world stabilization against complex, dynamic slung loads.
\end{itemize}

Future work will focus on extending this adaptive framework to handle highly turbulent aerodynamic wind gusts. Addressing such severe aerodynamic disturbances is essential for deploying these learned policies in more complex, real-world missions, such as autonomous ship-deck landings, where the vehicle must maintain precision within highly unsteady and turbulent ship wakes. Furthermore, we plan to validate the scalability of our approach by deploying the controller on larger quadrotor platforms to assess the RDP's performance across different mass, inertia, and actuation scales. Finally, a broader avenue for future research involves applying this methodology to alternative aerial configurations, including fixed-wing aircraft and hybrid vertical take-off and landing (VTOL) tailsitters, thereby expanding the impact of learned adaptive control to a wider range of challenging aerospace applications.

%% file: sec_08_acknowledgements.tex
\section{Acknowledgments}
This work is supported by the Office of Naval Research (ONR) under Grant Number N000142312404. The views and conclusions contained in this document are those of the authors and should not be interpreted as representing the official policies, either expressed or implied, of the Navy or the U.S. Government. The U.S. Government is authorized to reproduce and distribute reprints for Government purposes, notwithstanding any copyright notation herein.